\documentclass[10pt,twocolumn,letterpaper]{article}

\usepackage{cvpr}
\usepackage{times}
\usepackage{graphicx}

 \usepackage{url}            
 \usepackage{booktabs}       
 \usepackage{amsfonts}       
 \usepackage{nicefrac}       
 \usepackage{microtype}      
 \usepackage{multirow}
 \usepackage{color}
 \usepackage{amsmath,amssymb} 
 \usepackage[export]{adjustbox}
 \usepackage{algorithm}
 \usepackage{mathtools}
 \usepackage{algorithmic}
 \usepackage{wrapfig,booktabs}
 \usepackage{wrapfig}
 \usepackage{array}
\newcommand*\rot{\rotatebox{0}}

\usepackage[pagebackref=true,breaklinks=true,letterpaper=true,colorlinks,bookmarks=false]{hyperref}

\cvprfinalcopy 

\title{Efficient Parameter-free Clustering Using First Neighbor Relations}
\author{M. Saquib Sarfraz$^{1,2}$, Vivek Sharma$^1$, Rainer Stiefelhagen$^1$\\
$^1$Karlsruhe Institute of Technology\\ $^2$Daimler TSS, Germany\\
{\tt\small  \{firstname.lastname\}@kit.edu}}
\begin{document}
\pagenumbering{gobble}
\maketitle
\begin{abstract}
We present a new clustering method in the form of a single clustering equation that is able to directly discover groupings in the data. The main proposition is that the first neighbor of each sample is all one needs to discover large chains and finding the groups in the data. In contrast to most existing clustering algorithms our method does not require any hyper-parameters, distance thresholds and/or the need to specify the number of clusters. The proposed algorithm belongs to the family of hierarchical agglomerative methods. The technique has a very low computational overhead, is easily scalable and applicable to large practical problems. Evaluation on well known datasets from different domains ranging between 1077 and 8.1 million samples shows substantial performance gains when compared to the existing clustering techniques.  \href{https://github.com/ssarfraz/FINCH-Clustering}{[code release]}

\end{abstract}

\section{Introduction}
Discovering natural groupings in data is needed in virtually all areas of sciences. Despite half a decade of research in devising clustering techniques there is still no universal solution that 1.) can automatically determine the true (or close to true) clusters with high accuracy/purity; 2.) does not need hyper-parameters or a priori knowledge of data; 3.) generalizes across different data domains; and 4.) can scale easily to very large (millions of samples) data without requiring prohibitive computational resources. Clustering inherently builds on the notion of similarity. All clustering techniques strive to capture this notion of similarity by devising a criterion that may result in a defined local optimal solution for groupings in the data. Well known center-based methods (e.g., Kmeans) iteratively assign points to a chosen number of clusters based on their direct distance to the cluster center. Agglomerative clustering methods merge points based on predefined distance thresholds. More recent methods build similarity graphs (e.g., spectral clustering techniques) from the pairwise distances of the points and solve a graph partitioning problem by using these distances as edge weights and the sample points as nodes. All existing clustering techniques use some form of prior knowledge/assumptions on defining the similarity goal to obtain specific groupings in the data. This prior knowledge comes in the form of setting number of clusters in advance or setting distance thresholds or other hyper-parameters that render a user defined notion of similarity for obtaining groupings. These choices are subjective and must change when the underlying data distribution changes. This means that these parameters are not stable and need to be adjusted for each data set. This makes the clustering problem very hard as no standard solution can be used for different problems. 


In this paper, we describe an efficient and fully parameter-free unsupervised clustering algorithm that does not suffer from any of the aforementioned problems. By ``\textit{fully}", we mean that the algorithm does not require any user defined parameters such as similarity thresholds, or a predefined number of clusters, and that it does not need any a priori knowledge on the data distribution itself.  
The main premise of our proposed method is the discovery of an intriguing behavior of chaining large number of samples based on a simple observation of the first neighbor of each data point. Since the groupings so obtained do not depend on any predefined similarity thresholds or other specific objectives, the algorithm may have the potential of discovering natural clusters in the data. The proposed method has low computational overhead, is extremely fast, handles large data and provides meaningful groupings of the data with high purity.

\section{Related Work}
Books have been written to guide through a myriad of clustering techniques~\cite{clusterbook}. The representative algorithms can largely be classified in three directions, centroid/partitioning algorithms (e.g., Kmeans, Affinity Propagation), hierarchical agglomerative/divisive methods and methods that view clustering as a graph partitioning problem (e.g., spectral clustering methods).  For center-based clustering it is known that the Kmeans is sensitive to the selection of the initial
K centroids. The affinity propagation algorithm~\cite{ap} addresses this issue by viewing each sample as an exemplar and then an efficient message parsing mechanism is employed until a group of good exemplars and their corresponding clusters emerge. Such partition-based methods are also commonly approached by choosing an objective function and
then developing algorithms that approximately optimize that objective~\cite{raftery2006variable,kmeans,rcc}.
Spectral Clustering (SC) and its variants have gained popularity recently~\cite{von2007tutorial}.
Most spectral clustering algorithms need to compute the full similarity graph Laplacian matrix and have quadratic complexities, thus severely restricting the scalability of spectral clustering to large data sets. Some approximate algorithms have been proposed~\cite{yan2009fast, li2011time} to scale spectral methods. An important clustering direction has approached these spectral schemes by learning a sparse subspace where the data points are better separated, see Elhamifar and Vidal's Sparse Subspace Clustering (SSC)~\cite{elhamifar2013sparse}. The aim is to reduce the ambiguity in the sense of distances in high dimensional feature spaces. Recently many methods approach estimating such subspaces by also using low-rank constraints, see Vidal et al.~\cite{vidal2014low} and very recent work by Brbic and Kopriva~\cite{brbic2018multi}. 

In their remarkable classic work Jarvis \& Patrick~\cite{jp} bring forth the importance of shared nearest neighbors to define the similarity between points. The idea was rooted in the observation that two points are similar to the extent that their first k-neighbors match. Similarities so obtained are a better measure of distances between points than standard euclidean metrics. Using neighbors to define similarity between points has been used in partition-based, hierarchical and spectral clustering techniques~\cite{sc,bubeck2009nearest,ro}. 

Our paper closely relates to Hierarchical Agglomerative Clustering (HAC) methods, which have been studied extensively~\cite{reddy2013survey}.  The popularity of hierarchical clustering stems from the fact that it produces a clustering tree that provides meaningful ways to interpret data at different levels of granularity. For this reason, there is a lot of interest in the community to both develop and study theoretical properties of hierarchical clustering methods. Some of the recent works establish guarantees on widely used hierarchical algorithms for a natural objective function ~\cite{moseley2017approximation, cohen2017hierarchical}.
In agglomerative methods, each of the input sample points starts as a cluster. Then iteratively, pairs of similar clusters are merged according to some metric of similarity obtained via well studied linkage schemes. 
The most common linkage-based algorithms (single, average and complete-linkage) are often based on Kruskal’s minimum spanning tree algorithm~\cite{kruskal1956shortest}. The linkage methods merge two clusters based on the pairwise distances of the samples in them.
The linkage schemes can be better approached by an objective function that links clusters based on minimizing the total within cluster variance e.g., Ward~\cite{hac}. This approach generally produces better merges than the single or average linkage schemes. Dasgupta~\cite{dasgupta2016cost} recently proposed an objective function optimization on the similarity graph for hierarchical clustering to directly obtain an optimal tree. It initiated a line of work developing algorithms that explicitly optimize such objectives~\cite{roy2016hierarchical,charikar2017approximate, moseley2017approximation, cohen2017hierarchical}.

Recently clustering is also used in jointly learning a non-linear embedding of samples. This could be approached with deep learning based methods e.g., employing auto encoders to optimize and learn the features by using an existing clustering method~\cite{yang2016joint,xie2016unsupervised,guo2017improved,jiang2016variational}. These deep learning based approaches are primarily the feature representation learning schemes using an existing clustering method as a means of generating pseudo labels~\cite{caron2018deep} or as an objective for training the neural network.

Almost all of the existing clustering methods operate directly on the distance values of the samples. This makes it hard for these methods to scale to large data as they need to have access to the full distances stored as floats. Apart from this, all of the current clustering methods need some form of supervision/expert knowledge, from requiring to specify the number of clusters  to setting similarity thresholds or other algorithm specific hyper-parameters. Our proposal is a major shift from these methods in that we do not require any such prior knowledge and do not need to keep access to the full pairwise distance floats.

\section{The Proposed Clustering Method}

\begin{figure*}[t]
\centering
\begin{minipage}[b]{0.3\linewidth}
\centering
\centerline{\includegraphics[width=0.56\columnwidth]{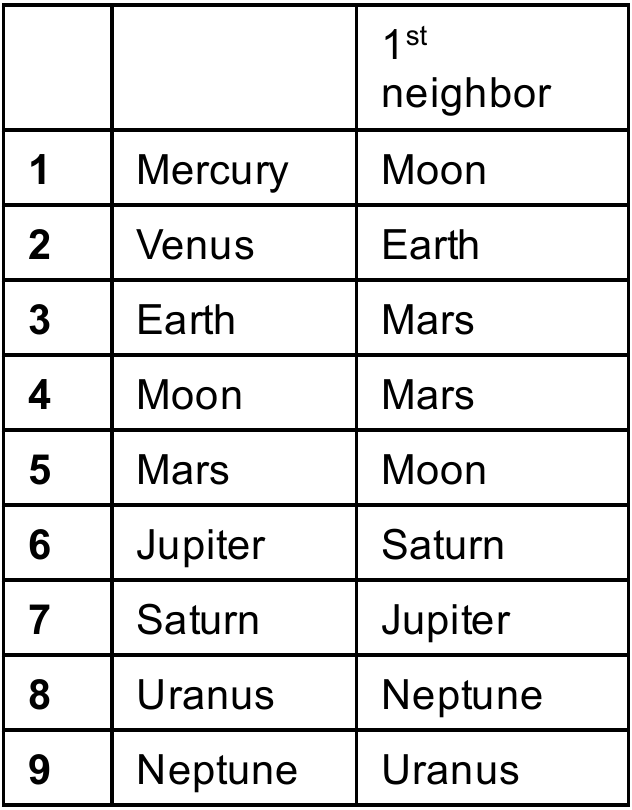}}
\centerline{ (a) }
\end{minipage}\hspace{0.1cm}
\begin{minipage}[b]{0.258\linewidth}
\centering
\centerline{\includegraphics[width=0.85\columnwidth]{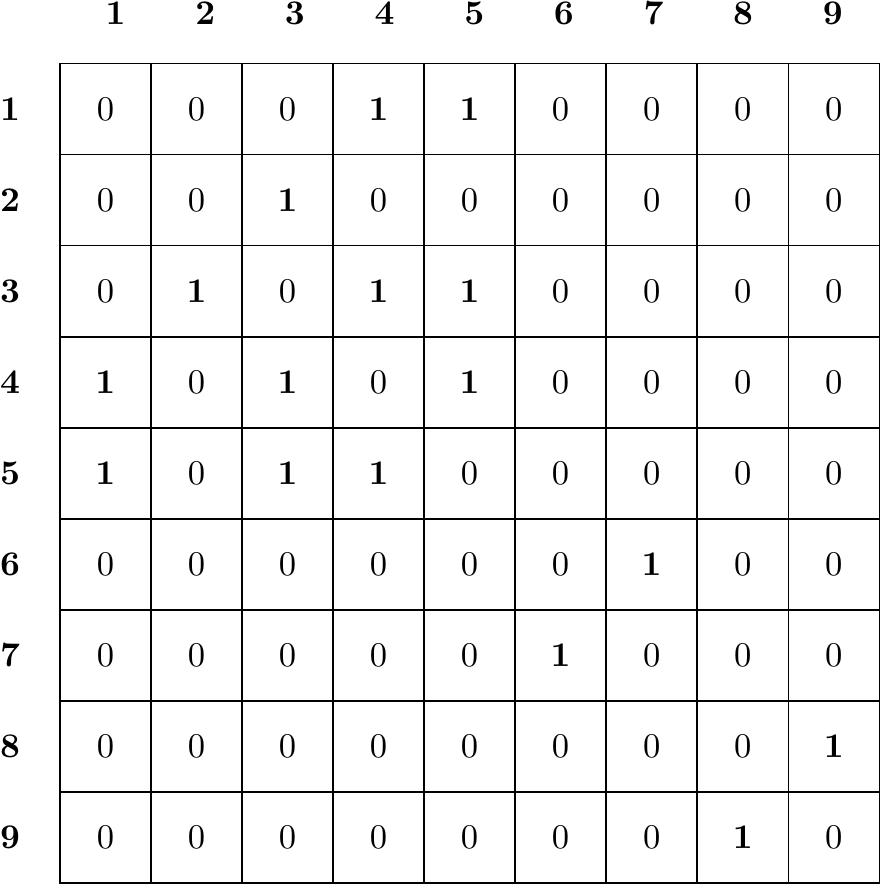}}
\centerline{ (b)}
\end{minipage} \hspace{0.1cm}
\begin{minipage}[b]{0.41\linewidth}
\centering
\centerline{\includegraphics[width=0.88\columnwidth]{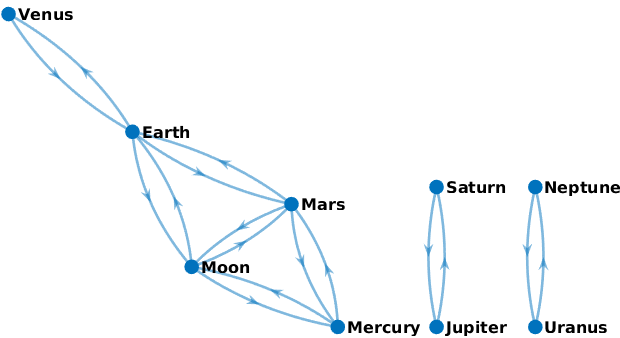}}
\centerline{ (c) }
\end{minipage}
\vspace{-0.10cm}
\caption{\small{\small{The clustering equation walk-through: clustering the planets in our solar system. (a) Planets and their first neighbors. (b) Adjacency link matrix by Eq.1. (c) Directed graph using the adjacency matrix of (b). FINCH discovered 3 clusters shown as directed graph's connected components. Each planet is represented by first 15 attributes described in \scriptsize{[\url{https://nssdc.gsfc.nasa.gov/planetary/factsheet/}}].}}} \label{fig:planets}
\vspace{-0.5cm}
\end{figure*}

 Historically, clustering methods obtain groupings of data by interpreting the direct distances between data points. Data that lies in high dimensional space has less informative measure of closeness in terms of these distances.
Methods that aim at describing uniform volumes of the hyperspace may fail because source samples are hardly uniformly distributed within the target manifold. 
On the other hand, semantic relations (i.e., who is your best friend/ or friend of a friend) may be impervious to this as they rely on indirect relations rather than exact distances. Our proposal is intuitively related to such semantic relations for discovering the groupings in the data.
We observed that the very first neighbor of each point is a sufficient statistic to discover linking chains in the data. Thus, merging the data into clusters can be achieved without the need to maintain a full distance matrix between all the data points. Also, using this approach, no thresholds or other hyper-parameters have to be set.
We capture this observation in \textit{the proposed clustering equation} and then describe the full algorithm in detail. 

\subsection{The Clustering Equation}

Given the integer indices of the first neighbor of each data point, we directly define an adjacency link matrix 
\begin{equation}
A(i,j) =
\begin{dcases*}
   1                         & if $j= \kappa_i^1$ or $\kappa_j^1=i$ or $\kappa_i^1=\kappa_j^1$ \\
   0                         & otherwise
\end{dcases*} \\
\label{eq:1}
\end{equation}

where $\kappa_i^1$ symbolizes the first neighbor of point $i$. The adjacency matrix links each point $i$ to its first neighbor via $j= \kappa_i^1$, enforces symmetry via $\kappa_j^1=i$ and links points $(i,j)$ that have the same neighbor with $\kappa_i^1=\kappa_j^1$. Equation~\ref{eq:1} returns a symmetric sparse matrix directly specifying a graph who's connected components are the clusters. One can see, in its striking simplicity, the clustering equation delivers clusters in the data without relying on any thresholds or further analysis. The connected components can be obtained efficiently from the adjacency matrix by using a directed or undirected graph $G=(V,E)$, where $V$ is the set of nodes (the points to be clustered) and the edges $E$ connect the nodes $A(i,j)=1$. Intuitively the conditions in the Equation~\ref{eq:1} are combining 1-nearest neighbor (1-nn) and shared nearest neighbour (SNN) graphs. Note, however, that the proposed clustering equation directly provides clusters without solving a graph segmentation problem. More precisely the adjacency matrix obtained from the proposed clustering equation has absolute links. We do not need to use any distance values as edge weights and require solving a graph partitioning. This is what makes it unique as by just using the integer indices of first neighbors, Equation~\ref{eq:1} specifies and discover the semantic relations and directly delivers the clusters. Computing the adjacency matrix is also computationally very efficient since it can easily be obtained by simple sparse matrix multiplication and addition operation.

To understand the mechanics of Equation~\ref{eq:1} and see how it chains large numbers of samples in the data, let's first look at a small tractable example of clustering the planets in our solar system. We can cluster the planets in the space of their known measurements e.g., mass, diameter, gravity, density, length of day, orbital period and orbital velocity etc. We describe each planet by the first 15 measurements taken from NASA's planetary fact sheet. The first neighbor of each planet in this space is the one with the minimum distance, obtained by computing pairwise euclidean distance. Figure~\ref{fig:planets} (a) shows the 9 samples (8 planets and moon) and their first neighbors. With this information one can now form the $9\times9$ adjacency matrix (see Fig.~\ref{fig:planets} (b)) according the Equation~\ref{eq:1}. An important observation is that not all first neighbors are symmetric, e.g., Earth is the first neighbor of Venus but Venus is not the first neighbor of Earth. This is also the basis of why these neighbors can form chains. Note how each of the adjacency conditions results in linking the planets. For instance, the adjacency condition $j= \kappa_i^1$ simply connects all planets to their first neighbors. The condition $\kappa_i^1=\kappa_j^1$ connects Mercury-Mars and Earth-Moon together because they have same first neighbor. The condition $\kappa_j^1=i$ forces symmetry and bridges the missing links in the chains, e.g., Earth is connected to Venus via this condition. Figure~\ref{fig:planets} (c) shows the directed graph of this adjacency matrix. Note how five out of nine planets are chained together directly while symmetric neighbors formed two separate clusters. This short walk-through explains the mechanics of Equation~\ref{eq:1} which has discovered three clusters.  Interestingly astronomers also distinguish these planets into three groups at a fine level, rocky planets (Mercury, Venus, Earth, Moon, and Mars) in one group,  gas planets (Jupiter, Saturn) being similar in terms of larger size with metallic cores, and ice giants (Uranus, Neptune) are grouped together because of similar atmospheres with rocky cores.




\subsection{Proposed Hierarchical Clustering} \label{subsec:algo}
The question whether the discovered clusters are indeed the true groupings in the data has no obvious answer. It is because the notion of clusters one considers true are subjective opinions of the observer. The problem of finding ground-truth clustering has been well studied in Balcan \etal~\cite{balcan2008discriminative}, where they show that having a list of partitions or a hierarchy instead of a single flat partition should be preferred. In such a set of groupings, they show that single or average-linkage algorithms are known to provably recover the ground-truth clustering under some properties of a similarity function. 

Equation~\ref{eq:1} delivers a flat partition of data into some automatically discovered clusters. Following it up recursively in an agglomerative fashion may provide further successive groupings of this partition capturing the underlying data structure at different level of granularities. 
Because Equation~\ref{eq:1} may chain large numbers of samples quickly, we show that in only a few recursions a meaningful small set of partitions is obtained with a very high likelihood of recovering the exact ground-truth clustering or a partition very close to it. Since by just using the \textbf{F}irst \textbf{I}nteger \textbf{N}eighbor indices we can produce a \textbf{C}lustering \textbf{H}ierarchy, we term our algorithm as (FINCH).

\begin{algorithm}[t]
\caption{Proposed Algorithm}\label{algo:the_alg}
\begin{algorithmic}[1]
\STATE \textbf{Input:} Sample set $S=\{1,2,\cdots,N\}$, $S \in \mathbb{R}^{ N \times d}$, where $N$ is total number of samples and each sample point is represented by $d$ attributes or feature dimensions.
\STATE \textbf{Output:} Set of Partitions $\mathbf{\mathcal{L}}=\{\Gamma_1,\Gamma_2,\cdots,\Gamma_L\}$ where each partition $\Gamma_i=\{C_1,C_2,\cdots,C_{\Gamma_i} \vert  C_{\Gamma_i} > C_{\Gamma_i+1} \forall i\in \mathcal{L} \}$ is a valid clustering of $S$. 
\STATE \textbf{The FINCH Algorithm:}
\STATE Compute first neighbors integer vector $\kappa^1 \in \mathbb{R}^{ N \times 1} $ via exact distance or fast approximate nearest neighbor methods.
\STATE  Given $\kappa^1$ get first partition $\Gamma_1$ with $C_{\Gamma_1}$ clusters via Equation~\ref{eq:1}. $C_{\Gamma_1}$  is the total number of clusters in partition $\Gamma_1$.
\WHILE{there are at least two clusters in $\Gamma_i$}
\STATE  Given input data $S$ and its partition $\Gamma_i$, compute cluster means (average of all data vectors in that cluster). Prepare new data matrix $M=\{1,2,\cdots,C_{\Gamma_i}\}$, where   $\mathbb{M}^{ C_{\Gamma_i} \times d}$.
\STATE Compute first neighbors integer vector $\kappa^1 \in \mathbb{R}^{ C_{\Gamma_i} \times 1} $ of points in $M$.
\STATE Given $\kappa^1$ get partition $\Gamma_M$ of $\Gamma_i$ via Equation~\ref{eq:1}, where $\Gamma_M\supseteq\Gamma_i$
\IF {$\Gamma_M$ has one cluster}
\STATE break
\ELSE
\STATE Update cluster labels in $\Gamma_i:\Gamma_M\to\Gamma_i$
\ENDIF 
\ENDWHILE
\end{algorithmic}
\end{algorithm}

\begin{algorithm}[t]
\caption{Required Number of Clusters Mode}\label{algo:the_alg2} 
\begin{algorithmic}[1]
\STATE \textbf{Input:} Sample set $S=\{1,2,\cdots,N\}$, $S \in \mathbb{R}^{ N \times d}$ and a partition $\Gamma_i$ from the output of Algorithm~1.
\STATE \textbf{Output:} Partition $\Gamma_r$ with required number of clusters.
\FOR{steps = $C_{\Gamma_i}$ - $C_{\Gamma_r}$}
\STATE  Given input data $S$ and its partition $\Gamma_i$, compute cluster means (average of all data vectors in that cluster). Prepare new data matrix $M=\{1,2,\cdots,C_{\Gamma_i}\}$, where   $\mathbb{M}^{ C_{\Gamma_i} \times d}$.
\STATE  Compute first neighbors integer vector $\kappa^1 \in \mathbb{R}^{ C_{\Gamma_i} \times 1} $ of points in $M$.
\STATE  Given $\kappa^1$ compute Adjacency Matrix of Equation~1
\STATE  $\forall A(i,j)=1$ keep one symmetric link $A(i,j)$ which has the minimum distance $d(i,j)$ and set all others to zero. 
\STATE  Update cluster labels in $\Gamma_i$: Merge corresponding $(i,j)$ clusters in $\Gamma_i$
\ENDFOR
\end{algorithmic}
\end{algorithm}

\noindent\textbf{The FINCH Algorithm.} The main flow of the proposed algorithm is straightforward. After computing the first partition, we want to merge these clusters recursively. For this, Equation~\ref{eq:1} requires the first neighbor of each of these clusters. Finding these neighbors requires computing distances between clusters. This is also related to all the linkage methods, e.g., average-linkage in hierarchical agglomerative clustering use the average of pairwise distances between all points in the two clusters. Following this, however, is computationally very expensive with a complexity of $\mathcal{O}(N^2log(N))$. It thus can not easily scale to millions of samples, and it requires to keep the full distance matrix in the memory. With our method, for large scale data, we do not need to compute the full pairwise distance matrix as we can obtain first neighbors via fast approximate nearest neighbor methods (such as k-d tree). We, instead, average the data samples in each cluster and use these mean vectors to compute the first neighbor. This simplifies the computation and keeps the complexity to $\mathcal{O}(Nlog(N))$ 
as we merge these clusters at each recursion. In contrast to standard HAC algorithms, our clustering equation directly provides a meaningful partition of the data at each iteration. It provides a hierarchical structure as a small set of partitions where each successive partition is a superset of the preceding partitions. The resulting set of partitions provides a fine to coarse view on the discovered groupings of data. The complete algorithmic steps are described in Algorithm~\ref{algo:the_alg}.

The quality of the outlined procedure is demonstrated in the experimental section on diverse data of different sizes. We show that it results in partitions that are very close to the ground-truth clusters. For some applications, it is often a requirement to obtain a specific number of clusters of the data. Since our algorithm returns a hierarchy tree, we can use simple mechanisms to refine a close partition by one merge at a time to provide required clusters as well. However, the required number of clusters are obviously upper-bounded by the size of the first partition, FINCH returns. For completeness, we outline a simple procedure in Algorithm~2.

\begin{figure*}[t]
\centering
\resizebox{14cm}{!} {
    \begin{minipage}[b]{0.19\textwidth}
    \centering
    \centerline{\includegraphics[width=1\columnwidth, frame]{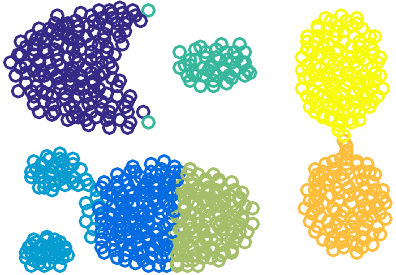}}
    \centerline{ Kmeans: 88\% }
    \end{minipage}
    \begin{minipage}[b]{0.19\linewidth}
    \centering
    \centerline{\includegraphics[width=1\columnwidth, frame]{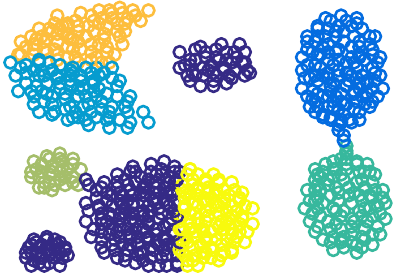}}
    \centerline{ Spectral: 81\% }
    \end{minipage}
    \begin{minipage}[b]{0.19\linewidth}
    \centering
    \centerline{\includegraphics[width=1\columnwidth, frame]{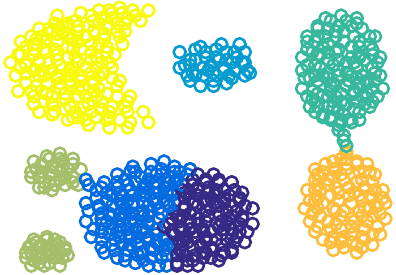}}
    \centerline{ HAC: 92\% }
    \end{minipage}
    \begin{minipage}[b]{0.19\linewidth}
    \centering    
    \centerline{\includegraphics[width=1\columnwidth, frame]{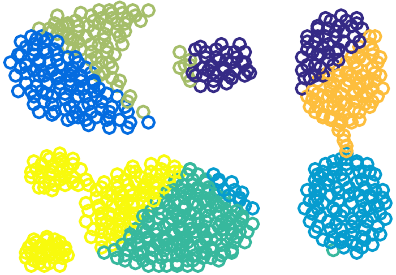}}
    \centerline{ SSC: 74\% }
    \end{minipage}
    \begin{minipage}[b]{0.19\linewidth}
    \centering
    \centerline{\includegraphics[width=1\columnwidth, frame]{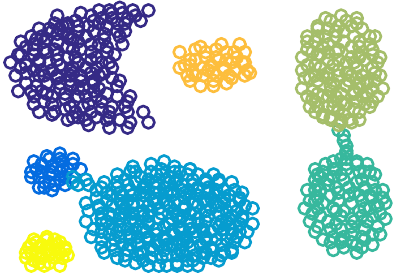}}
    \centerline{ \textbf{FINCH: 98\% }}
    \end{minipage}}
    \qquad \qquad
    \resizebox{14cm}{!} {
    \begin{minipage}[b]{0.19\linewidth}
    \centering
    \centerline{\includegraphics[width=1\columnwidth, frame]{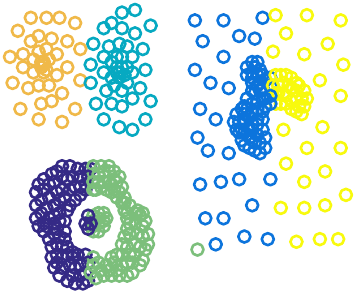}}
    \centerline{ Kmeans~\cite{kmeans}: 72\% }
    \end{minipage}
    \begin{minipage}[b]{0.19\linewidth}
    \centering
    \centerline{\includegraphics[width=1\columnwidth, frame]{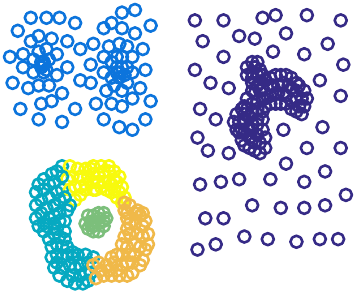}}
    \centerline{ Spectral~\cite{sc}: 75\% }
    \end{minipage}
    \begin{minipage}[b]{0.19\linewidth}
    \centering
    \centerline{\includegraphics[width=1\columnwidth, frame]{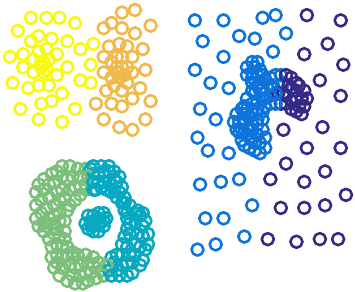}}
    \centerline{ HAC~\cite{hac}: 73\% }
    \end{minipage}
    \begin{minipage}[b]{0.19\linewidth}
    \centering
    \centerline{\includegraphics[width=1\columnwidth, frame]{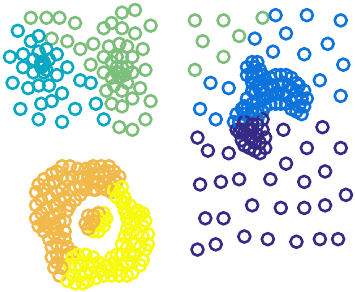}}
    \centerline{ SSC~\cite{elhamifar2013sparse}: 67\% }
    \end{minipage}
    \begin{minipage}[b]{0.19\linewidth}
    \centering
    \centerline{\includegraphics[width=1\columnwidth, frame]{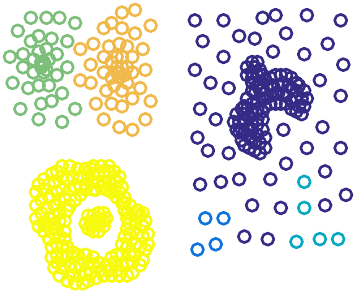}}
    \centerline{ \textbf{FINCH: 85\% }}
    \end{minipage}}
	\caption{Visualization of Aggregation~\cite{gionis2007clustering} 7 clusters (top) and Gestalt~\cite{zahn1970graph} 6 clusters (bottom) with NMI score for each method.}
	\label{fig:2dplot}
	\vspace{-0.50cm}
\end{figure*}


For better clarity, we demonstrate the quality of the outlined FINCH algorithm in Fig.~\ref{fig:2dplot} using the classic 2D Gestalt~\cite{zahn1970graph} and Aggregation~\cite{gionis2007clustering} data that represent well understood clustering problems and provide a good qualitative look into the performance of clustering methods. After running FINCH Algo~\ref{algo:the_alg} we obtained a set of clustering partitions. Here, to evaluate FINCH on the true clusters, we can take a larger partition than the true clusters and refine it with Algo~\ref{algo:the_alg2}. For example, on Gestalt data with 399 2D points, Algo~\ref{algo:the_alg} provides a total of 4 partitions with \{91, 23, 5 and 2\} clusters. We use Algo~\ref{algo:the_alg2} on the 23 cluster partition to evaluate at ground truth 6 clusters. Clearly, we can observe that FINCH maintains the merges quite well and clusters these data better than the considered baselines.


\section{Experiments}
We demonstrate FINCH on challenging datasets which cover domains such as biological data, text, digits, faces and objects. We first introduce the datasets and clustering metric, followed by a thorough comparison of the proposed method to algorithms that estimate the clusters automatically using hyper-parameter settings, and also to algorithms that require \#clusters as input. We also compare FINCH with state-of-the-art deep clustering methods in section~\ref{sec:deepclust}.

\noindent\textbf{Datasets.} 
The datasets are summarized in Table~\ref{table:datasets}. The dimensionality of the data in the different datasets varies from 77 to 4096. \texttt{Mice Protein}~\cite{miceprotein} consists of the expression levels of 77 proteins measured in eight classes of control and trisomic genotypes, available from~\cite{rcc}.
\texttt{REUTERS}~\cite{reuters} consists of 800000 Reuters newswire articles. We use a sampled subset of 10000 articles available from~\cite{xie2016unsupervised}, categorized in four classes.  The TF-IDF features are of 2000 dimensions. \texttt{STL-10}~\cite{stl10} is an image recognition dataset for unsupervised learning, consisting of 10 classes with 1300 examples each. \texttt{BBTs01}~(season 1, episodes 1 to 6) and \texttt{BFs05}~(season 5, episodes 1 to 6) are challenging video face identification/clustering datasets. They are sitcoms with small cast list, for BBTs01: 5 main casts, and BFs05: 6 main casts. Data for BBTs01 and BFs05 are provided by~\cite{baeuml2013}. \texttt{MNIST}~\cite{mnist}:  we use three variants of MNIST handwritten digits, they consist of: 10K testset~(\texttt{MNIST\_10k}), 70K (train+test)~(\texttt{MNIST\_70k}), and 8.1M~\cite{mnist8m}~(\texttt{MNIST\_8M}) samples categorized into 10 classes.  We use CNN features for STL-10, BBTs01, BFs05, and both CNN features and pixels for MNIST datasets. More details on the features used and datasets settings can be found in the appendix .



\begin{table*}[htb]
\begin{center}
\resizebox{14cm}{!} {
\begin{tabular}{l |cc|ccc|ccc}
\toprule
  					& Mice Protein & REUTERS & STL-10 & BBTs01 & BFs05   & \multicolumn{3}{c}{MNIST} \\
\midrule
Type  				& Biological  &	Text& Objects &	\multicolumn{2}{c|}{Faces} &\multicolumn{3}{c}{Digits}  \\
\#C  			    &8 &	4&	10&	5&	6 &	\multicolumn{3}{c}{10}  \\
\#S  			&1077 &	10k&	13k&	199346 &206254& 	10k&	  70k & 8.1M  \\
Dim.  				&77 &	2000&	2048&	2048&	2048&   4096/784&	4096/784 & 256/784   \\
LC/SC~(\%)  			&13.88/9.72 &43.12/8.14	&10/10	&33.17/0.63 &39.98/0.61	& 11.35/8.92& 11.25/9.01 &11.23/9.03  \\
\bottomrule
\end{tabular}}
\end{center}
\vspace{-0.25cm}
\caption{\small{Datasets used in experiments. \#S denotes the number of samples, \#C denotes the true number classes/clusters, and largest class (LC) / smallest class (SC) is the class balance percent of the given data.}}\label{table:datasets}
\vspace{-0.15cm}
\end{table*}

\noindent\textbf{Evaluation Metrics.} We use the most common clustering evaluation metric, Normalized Mutual Information (NMI),
and the unsupervised clustering accuracy (ACC) as the metric to evaluate the quality of clustering. ACC is also widely used \cite{xie2016unsupervised,guo2017improved,jiang2016variational} and computed as
$\genfrac{}{}{0pt}{}{max}{m} \frac{\sum_{i=1}^{n} \mathbf{1}\{l_{i}=m(c_{i})\}}{n}$, where $l_{i}$ is the ground truth label, $c_{i}$ is the cluster assignment obtained by the method, and $m$ ranges in the all possible one-to-one mappings between clusters and labels. 

\noindent\textbf{Baselines.} We compare FINCH to existing clustering methods (both classic and recent proposals) that covers the whole spectrum of representative clustering directions. We include 11 baselines categorized in two variants of clustering algorithms: (1) algorithms that estimate the number of clusters automatically - given some input hyper-parameter/threshold settings. These algorithms include Affinity Propagation~(AP)~\cite{ap}, Jarvis-Patrick~(JP)~\cite{jp}, Rank-Order~(RO)~\cite{ro}, and the recently proposed Robust Continuous Clustering~(RCC)~\cite{rcc}; and (2) algorithms that require the number of clusters as input. These algorithms are: \textit{Kmeans++}~(Kmeans)~\cite{kmeans}, Birch~(BR)~\cite{br}, Spectral~(SC)~\cite{sc}, Sparse Subspace Clustering~(SSC)~\cite{elhamifar2013sparse}, Hierarchical Agglomerative Clustering~(HAC\_Ward) with ward linkage~\cite{hac}, and HAC with average linkage~(HAC\_Avg), and very recent method Multi-view Low-rank Sparse Subspace Clustering~(MV-LRSSC)~\cite{brbic2018multi}. The parameter settings for the baselines are provided in the \textit{appendix}.

\begin{table*}[htb]
\small
\tabcolsep=0.2cm
\begin{center}
\resizebox{17cm}{!} {
\begin{tabular}{  l | c |      c c c c |  c   c   c  c c c c | c c}
\toprule
   &  \multicolumn{12}{c|}{NMI Scores}&  &\\
   \cline{2-13}
   &  \multicolumn{5}{c|}{\color{blue}{Algorithms that estimate \#C automatically}}    & \multicolumn{7}{c|}{\color{blue}{@FINCH estimated \#C}}&  &\\
\cline{2-13}
 Dataset   & \textbf{\rot{FINCH}} 	 & \rot{AP} & \rot{JP} & \rot{RO}  & \rot{RCC}& \rot{Kmeans} 	& \rot{BR} & \rot{SC} & \rot{HAC\_Ward} & \rot{HAC\_Avg} & \rot{SSC} & \rot{MV-LRSSC}  & \rot{True \#C} & \rot{\#S}\\
\midrule
\midrule
Mice Protein 	&\textbf{51.64} & 59.10 & 55.99 &1.75 &\textbf{65.92} &42.66 & 40.39 & 55.13  &51.35 &37.65 & 41.94& 51.31\\
Estim. \#C  	  & \textbf{8}  &67  &30 	&2 &38    &  \multicolumn{7}{c|}{8}  & 8 & 1077\\
\midrule
REUTERS~\cite{}&   \textbf{44.88} & 36.23 & 22.97 & 36.76& 28.32 & 41.75 &38.77 & 7.97& 38.40 & 12.38  & 3.19 &41.27 \\
Estim. \#C  	& \textbf{4}	& 1073 &  1656	&  9937& 358&  \multicolumn{7}{c|}{4} & 4 & 10k\\
\midrule
STL-10~\cite{}  &    \textbf{85.05} & 57.18 &51.70 &33.37 & 81.56 & 85.59 & 80.9 & 72.62 &80.9 & 52.57 & 81.25& 74.44 \\
Estim. \#C  	&    \textbf{10} & 589 &  4780&  4358&14&  \multicolumn{7}{c|}{10}   & 10& 13k\\
\midrule
MNIST\_10k~\cite{} 	&  \textbf{97.55} &69.97 & 35.97 & 49.87 & 77.74 & 81.92 & 80.78 & 97.43 & 89.05 &63.86 & 96.63 & 93.67  \\
Estim. \#C  	&  \textbf{10}		& 116 &  513&  9950& 149&  \multicolumn{7}{c|}{10} & 10 & 10k\\
\midrule
MNIST\_70k~\cite{} 	&\textbf{98.84} & $-$ &24.20 & 4.01 & 86.59 & 81.02 & 84.50 &98.77 &87.61 & 47.08 &$-$ &$-$ \\
Estim. \#C  		&  \textbf{10}	&$-$  & 5722 &531 & 120&  \multicolumn{7}{c|}{10} & 10 & 70k\\
\bottomrule
\end{tabular} }
\end{center}
\vspace{-0.25cm}
\caption{\small{Small-scale clustering results of FINCH with nearest neighbors obtained using exact distances.  We compare FINCH against algorithms that estimates the clusters automatically - given input hyper-parameters/thresholds, and the algorithms that requires the \#clusters as input. For algorithms that require the number of clusters as input, we use the \#clusters estimated by FINCH. $-$ denotes OUT\_OF\_MEMORY.}} \label{table:all_dataset} 
\vspace{-0.5cm}
\end{table*}

\subsection{Comparison with baselines}
\noindent
\textbf{Comparison on Small-scale datasets:} In this section, we consider clustering datasets upto 70k samples. Considering the size of BBTs01 ($\sim$199k) samples, the full distance matrix  takes up approximately 148 GB RAM. The memory consumption of different algorithms is not linear rather exponential, and the number of samples and feature dimensions parameters negatively influence their time efficiency and computational cost. For this reason, we separate algorithms that need to store quadratic memory usage. The algorithms that need to compute the full distance matrix are evaluated in this section: small scale ($\le$70k: $\sim$36.5 GB), while large scale ($\ge$199k: $\sim$148 GB) are evaluated separately.

In Table~\ref{table:all_dataset}, we compare the performance of FINCH with the current clustering methods.  Results on datasets: MICE Protein, REUTERS, STL-10, MNIST\_10k and MNIST\_70k are reported in Table~\ref{table:all_dataset} using NMI-measure. 
We compare FINCH against the algorithms that requires the number of cluster as input: Kmeans, BR, SC, HAC, SSC and MV-LRSSC. To have a fair time comparison with these algorithms we also compute the full pairwise distance matrix for obtaining the first neighbours. To demonstrate the quality of merges FINCH made, we use the FINCH estimated clusters as input to these algorithms. On these datasets, FINCH has estimated the ground truth clusters as one of its partition, see Fig~\ref{fig:CUCAplots} and discussion in section~\ref{sec:discussion}. 

For comparison with algorithms that provides a fixed flat partition given some hyperparamters (AP, JP, RO and RCC), we can not directly compare FINCH as it provides a hierarchy. Here we follow the previous approaches~\cite{otto2018clustering,rcc,shah2018deep} that tend to compare on the basis of NMI measure only, and not on the basis of estimated number of clusters. In Table~\ref{table:all_dataset}, following the same procedure, we simply evaluate all the algorithms at the  respective author's best parameter setup for each method, and report their results attained. We observe that not only FINCH finds a meaningful partition of the data it also consistently achieves high performance on most of these datasets. \\
\textbf{Comparison on Large-scale datasets:} As FINCH only requires the first neighbor indices, for large scale datasets we obtain the first nearest neighbor using the randomized k-d tree algorithm from~\cite{muja2014scalable}, thus avoiding the expensive $\mathcal{O}(N^2)$ distance matrix computation cost, and quadratic memory storage. For example, computing the full distance matrix for single precision MNIST\_8M requires 244416.3 GB RAM. 

Among all our considered baselines, only Kmeans and RCC are able to scale. We compare FINCH against Kmeans and RCC on BBTs01 and BFs05 datasets. On the million scale (8.1 million samples) MNIST\_8M datasets, we were only able to run Kmeans.

\begin{table}[t]
\small
\tabcolsep=0.2cm
\begin{center}
\resizebox{6cm}{!} {
\begin{tabular}{   l | cc |  c    }
\toprule
    & \multicolumn{3}{c}{NMI}  \\ 
\cline{2-4}
  Dataset   & \textbf{{FINCH}} & RCC& Kmeans \\
\midrule
\midrule
BBTs01  & \textbf{89.79}& 2.56 & 71.82 \\
Estim.~\#C& 6&  7& 6\\
\midrule
BBTs01 (@True~\#C=5) & \textbf{91.57}& $-$ & 83.39  \\
\midrule
\midrule
BFs05 & \textbf{82.46} & 46.70 & 71.85 \\ 
Estim.~\#C & 7&  523& 7 \\
\midrule
BFs05 (@True~\#C=6) & \textbf{83.64} &  $-$ & 76.15  \\ 
\bottomrule
\end{tabular}}
\end{center}
\vspace{-0.25cm}
\caption{BBTs01 and BFs05~($\sim$200k).} \label{table:bbt_bf}
\vspace{-0.2cm}
\end{table}

\begin{table}[t]
\small
\tabcolsep=0.1cm
\begin{center} 
\resizebox{7cm}{!} {
\begin{tabular}{   l | cc    }
\toprule
    & \multicolumn{2}{c}{NMI}  \\ 
\cline{2-3}
  Dataset    & \textbf{\rot{FINCH}} & \rot{Kmeans}\\
\midrule
\midrule
MNIST\_8M\_CNN (@Estim.~\#C=13) & \textbf{96.55} & 93.33\\
MNIST\_8M\_CNN (@True~\#C=10) & \textbf{99.54} & 97.39\\
\midrule
MNIST\_8M\_PIXELS (@Estim.~\#C=11) & \textbf{46.49} & 40.17\\ 
MNIST\_8M\_PIXELS (@True~\#C=10)  & \textbf{63.84} & 37.26\\ 
\bottomrule
\end{tabular}}
\end{center}
\vspace{-0.25cm}
\caption{MNIST\_8M~(8.1M).} \label{table:mnist_8m}
\vspace{-0.5cm}
\end{table}

\begin{table*}[t]
\small
\tabcolsep=0.04cm
\begin{center}
\resizebox{14cm}{!} {
\begin{tabular}{  l | l c |     l    | c c c c c   c   c c c c c }
\toprule
 Dataset & \#S & Dimen. 	& \textbf{\rot{FINCH}}    & \rot{Kmeans} & \rot{SC} & \rot{HAC\_Ward} & \rot{HAC\_Avg} 	 & \rot{AP} & \rot{JP} & \rot{RO} & \rot{BR} & \rot{RCC} & \rot{SSC} & \rot{MV-LRSSC} \\
\midrule
Mice Protein  	&1077	&77		& \textbf{37ms}	 &115ms &220ms  &40ms  &668ms &700ms &00:01 &00:02 &90ms &84ms 	&00:08 & 00:02\\
REUTERS  		&10k	&2000	&\textbf{00:05}  &00:18 &05:54	&00:31 &00:43 &01:53 &40:25 &00:14 &01:32 &37:25&\color{red}{01:27:36} & 52:53 \\
STL-10 			&13k 	&2048	&\textbf{00:03}  &00:19	&08:03  &00:42 &01:10 &02:42 &57:49 &00:07 &02:25 &15:11& \color{red}{02:41:14} & \color{red}{02:04:52}\\
MNIST\_10k  	&10k	&4096	&\textbf{00:10}	 &00:19 &02:39	&01:05 &01:31 &02:23 &44:20 &00:12 &03:06 &13:41& \color{red}{01:35:25} & 38:42 \\
MNIST\_70k  	&70k	&4096	&\textbf{00:54}  &02:19 &58:45 &29:28   &30:17 &$-$ &\color{red}{60:09:17} &05:22 &\color{red}{02:20:44} &\color{red}{05:53:43} & $-$ & $-$\\
\midrule
BBTs01  		&199346	&2048	&\textbf{01:06} &02:17 &$-$  &$-$ &$-$ &$-$  &$-$ &$-$ &$-$&\color{red}{00:38:11} &$-$ &$-$\\
BFs05   		&206254	&2048	&\textbf{01:09} &01:33 &$-$  &$-$ &$-$ &$-$ &$-$ &$-$ &$-$&\color{red}{03:28:04} &$-$&$-$\\
MNIST\_8M  		&8.1M			&256	&\textbf{18:23} &56:41 &$-$	&$-$  &$-$ &$-$	&$-$ &$-$ &$-$&$-$&$-$&$-$\\
\midrule
Framework  		& 		&   	&Matlab	&Python	&Python	&Matlab &Matlab	&Python	&Matlab &C++ &Python &Python & Matlab & Matlab\\
\bottomrule
\end{tabular}}

\end{center}
\vspace{-0.25cm}
\caption{\small{Run-time comparison of FINCH with Kmeans, SC, HAC, AP, JP, RO, BR, RCC SSC, and MV-LRSSC. We report the run time in {\color{red}{HH:MM:SS}} and MM:SS. $-$ denotes OUT\_OF\_MEMORY. }} \label{table:time} 
\vspace{-0.5cm}
\end{table*}

For BBTs01 and BFs05, there exists a huge line of work, from exploiting video-level constraints~\cite{tc}, to dynamic clustering constraints via MRF~\cite{zhang2016joint}, and the most recent  link-based clustering~\cite{erdosclustering}. In contrast to these works, we use features from a pre-trained model on other datasets without any data specific transfer or considering any other video-level constraints for clustering. FINCH performs significantly better in comparison with the previously published methods on these datasets. These comparison results are available in the \textit{appendix}. Results in Table~\ref{table:bbt_bf} and run-time in Table~\ref{table:time} show that with approximate nearest neighbors, FINCH achieves the best run-time of the three methods and the best performance as well. A similar behavior can be observed in Table~\ref{table:mnist_8m} for very large scale MNIST\_8M datasets.

\subsection{Deep Clustering: Unsupervised Learning} \label{sec:deepclust}
Many recent methods propose to learn feature representations using a given clustering algorithm as objective, or as a means of providing weak labels/pseudo labels for training a deep model (usually an auto-encoder)~\cite{yang2016joint,xie2016unsupervised,guo2017improved,jiang2016variational, caron2018deep}.
FINCH lends itself naturally to this task since it provides a hierarchy of partitions containing automatically discovered natural groupings of data and one need not specify a user defined number of clusters to train the network. To demonstrate this and to be able to compare with deep clustering methods we follow a similar approach, as used recently in \cite{caron2018deep}, of using cluster labels to train a small network for learning an unsupervised feature embedding. We demonstrate this unsupervised learning on \texttt{MNIST\_70K\_PIXELS} (28x28 =784-dim), \texttt{REUTERS-10k} TF-IDF (2000-dim) and \texttt{STL-10} ResNet50 (2048-dim) features as input. Note that we use the same features that were used in previous methods to train their embedding, see Jiang \etal~\cite{jiang2016variational}.

We train a simple MLP with two hidden layers for this task in classification mode. Among the FINCH returned partitions the partition obtained at the first pass of the Equation~\ref{eq:1} or the one after it contains the largest number of clusters and also the purest since follow on recursions are merging these into smaller clusters. We use the estimated clusters of the second partition as the pseudo labels to train our network with softmax. We approach the training in steps, initializing training with the FINCH labels and re-clustering the last layer embedding after 20 epochs to update the labels. At each label update step we divide the learning rate by 10 and train for 60-100 epochs. For all experiments our network structure is fixed with the two hidden layers set to 512 neurons each. We train the network with minibatch SGD with batch size of 256 and initial learning rate of 0.01. The last layer 512-dimensional embedding is trained this way and used to report clustering performance at the ground truth clusters. As seen in Table~\ref{tab:deep_soa} FINCH has effectively trained an unsupervised embedding that clusters better in this learned space. Interestingly, on STL-10 for which we have ResNet50 features, FINCH directly clusters these base features with better accuracy as the compared methods achieve after training a deep model. After our FINCH driven deep clustering we are able to improve clustering performance on STL-10 to $\sim95\%$ improving by almost 10\% on the existing state-of-the-art deep clustering methods.

\begin{table}[t]
\small
\tabcolsep=0.1cm
\begin{center}
\resizebox{8.5cm}{!} {
\begin{tabular}{l|c c c }
\toprule
&   \multicolumn{3}{c}{Accuracy (ACC \%)}  \\
\cline{2-4}   
Method	& MNIST\_70k\_PIXELS	& REUTERS-10K 	& STL-10 \\
\midrule
GMM~\cite{jiang2016variational}		& 53.73	& 54.72			& 72.44 \\
AE+GMM~\cite{jiang2016variational}	& 82.18	& 70.13			& 79.83 \\
VAE+GMM~\cite{jiang2016variational}	& 72.94	& 69.56		 	& 78.86 \\
DEC~\cite{xie2016unsupervised}		& 84.30	& 72.17		 	& 80.62 \\
IDEC~\cite{guo2017improved}		& 88.06	& 76.05		 	& - \\
VaDE~\cite{jiang2016variational}	& \textbf{94.46}	& 79.83			& 84.45 \\
\midrule
FINCH on base features       & 74.00     & 66.14 & 85.28 \\  
DeepClustering With FINCH 	& 91.89		&	 \textbf{81.46}			&	\textbf{95.24}	\\ 
\bottomrule
\end{tabular}}
\end{center}
\vspace{-0.25cm}
\caption{Unsupervised Deep Clustering with FINCH: Comparison with state-of-the-art deep clustering methods at true clusters.} \label{tab:deep_soa}
\vspace{-0.48cm}
\end{table}

\subsection{Computational Advantage}
In Table~\ref{table:time}, we report the run-time of each algorithm for all the datasets. The corresponding timing is the complete time for the algorithm execution, that includes computing the pairwise distance between the samples, or obtaining the nearest neighbors using kd-tree, and the running time of each algorithm. We can observe that, FINCH achieves a remarkable time efficiency, and is not dominated by the number of samples, and/or the feature dimensions. Apart from time efficiency, FINCH is very memory efficient, as it only requires to keep the integer ($N\times1$) first neighbor indices and the data. The performance is measured on a workstation with an AMD Ryzen Threadripper 1950X 16-core processor with 192 (128+64 swap) GB RAM. For large scale datasets with more than 70k samples, most of the algorithms break, or demands for more than 192 GB RAM. FINCH memory requirement is, therefore, $\mathcal{O}(N)$ vs $\mathcal{O}(N^2)$. The computational complexity of FINCH is $\mathcal{O}(Nlog(N))$, whereas spectral methods are $\mathcal{O}(N^3)$ and hierarchical agglomerative linkage-based methods are $\mathcal{O}(N^2log(N))$.   

\section{Discussion} \label{sec:discussion}
\begin{figure*}[t]
 \centering
\includegraphics[width=1.7\columnwidth]{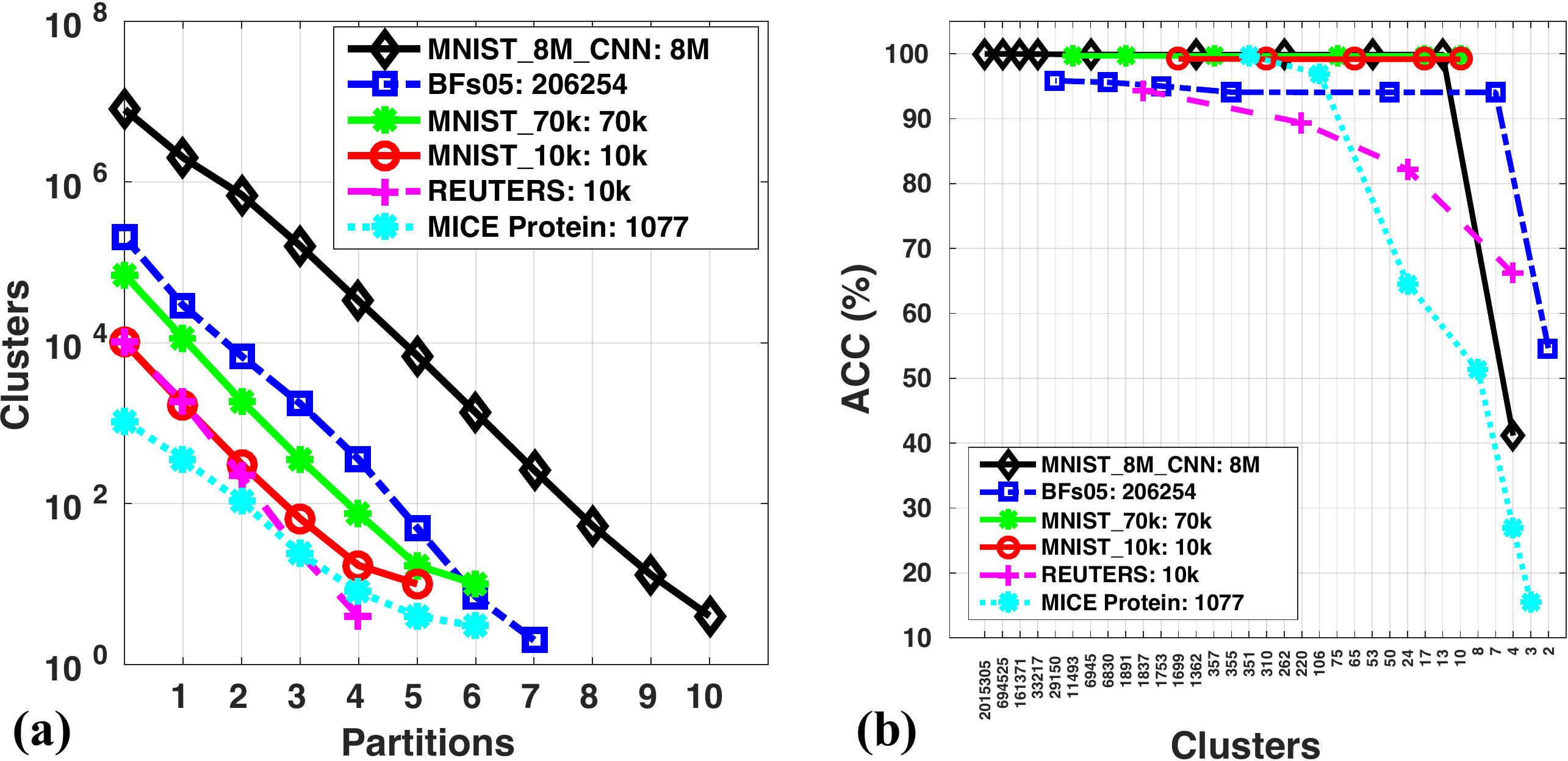} 
\caption{\small{\small{FINCH steps/partitions and clustering quality. (a) Algorithm convergence: clusters vs algorithm steps, each step produces a partition of data. (b) Quality of merges: number of cluster produced at each step and their corresponding purity/accuracy}}}\label{fig:CUCAplots}
\vspace{-0.5cm}
\end{figure*} 

We have extensively evaluated FINCH on a number of different data (image pixels, biological measurements, text frequency and CNN features) represented in 77 to 4096 dimensional feature space. Interestingly many existing popular clustering methods do not perform well even in the cases where the data has well separated distinct clusters in the feature space. 
The recursion scheme in our algorithm converges at an exponential rate in very few steps providing a valid partition of the data at each step. In Figure~\ref{fig:CUCAplots}~(a) we plot the clusters obtained at each step of the FINCH algorithm. One can see for the different data sizes (1077 samples to 8.1 million samples), our algorithm converges in 4-10 steps providing a valid partition at each. The accuracy of the formed clusters at each partition is depicted in Figure~\ref{fig:CUCAplots}~(b). One can assess the quality of discovered clusters and the successive merges by the fact that it maintains the accuracy quite well for very large merges through each step. Corresponding to Figure~\ref{fig:CUCAplots}~(a) the \textit{x-axis} of Figure~\ref{fig:CUCAplots}~(b)  depicts \#clusters obtained at each step of the algorithm for all of the plotted datasets. For example, for the smallest dataset with 1077 samples, FINCH delivers a total of 6 partitions in 6 steps of the run. From 1077 samples it comes down to 351 clusters in the first pass of Equation~\ref{eq:1} and then to 106, 24, 8, 4 and 3 clusters in the successive recursions, discovering the ground-truth clustering of 8 clusters at step 4 of the algorithm. One can interpret the corresponding cluster purity and the number of clusters at each step of FINCH for the other datasets in Figure~\ref{fig:CUCAplots} (b). In direct contrast to HAC linkage schemes which require $N-1$ steps to converge for $N$ samples, FINCH convergence (number of steps) does not depend on $N$ and is governed by Equation~\ref{eq:1}. It is very interesting to study this behavior and finding some theoretical bounds explaining this.
An interesting observation are the results on the MNIST\_10k and MNIST\_70k datasets in Table~\ref{table:all_dataset}.
Since these features have been learned on the data using the ground-truth labels, these are already well separated 10 clusters in the feature space. 
Despite this, none of the baseline algorithms (even the ones that require to specify the number of clusters)  performs as accurately. FINCH stops its recursive merges at step 5~(for MNIST\_10k) and step 6~(for MNIST\_70k) providing the exact 10 clusters with above 99\% accuracy. Note that this is the same classification accuracy as the trained CNN model provides on these features. Since the ground-truth clusters are well separated, Algorithm~\ref{algo:the_alg} exactly stops here because another pass of this 10-cluster partition merge them to 1 cluster. This shows that FINCH can recover well the global structure in the data.  We also include the number of steps and corresponding clusters, along with the accuracy, for each dataset in the \textit{appendix}. Note that, because of the definition of Equation~\ref{eq:1}, FINCH can not discover singletons (clusters with 1 sample). This is because we link each sample to its first neighbor without considering the distance between them. Singletons, therefore will always be paired to their nearest sample point. This sets the limit  of smallest cluster with size 2 for FINCH.\\
\indent Conclusively, we have presented an algorithm that shifts from the prevalent body of clustering methods that need to keep the pairwise distance matrix and require user defined hyper-parameters. 
It offers a unique and simple solution in the form of the clustering equation and an effective agglomerative merging procedure. The advantage it brings, in being fully parameter-free and easily scalable to \textit{large data} at a minimal computational expense, may prove useful for many applications. We have demonstrated one such application of unsupervised feature representation learning. Automatically discovering meaningful clusters in such a manner is needed in many areas of sciences where nothing is known about the structure of data. For example, it may have very exciting applications from discovering exotic particles to stars/galaxies based on their spectra.

{\small
\bibliographystyle{ieee}
\bibliography{CUCA}
}

\section*{Appendix}
\subsection*{Datasets Detail}
We have used CNN features for STL-10, BBT/Buffy and MNIST datasets. Specifically, we use ResNet50 pretrained on ImageNet for STL-10, and pretrained VGG2-Face ResNet50 model features~\cite{vgg2} for BBTs01, BFs05. On MNIST datasets, we show clustering results both on raw pixels and trained CNN features. Since the clustering algorithms performance varies with varying feature dimensions, we have used both 4096 and 256-dimesional features for MNIST to have this diversity of feature dimensions. We use \cite{jaderberg2014synthetic} CNN features (4096-dim) for MNIST\_10k and MNIST\_70k, while for MNIST\_8M we train a two layer MLP with 512 and 256 neurons with 60\% data used for training. Here, the last 256-dimensional layer is used as the feature extractor, \texttt{MNIST\_8M\_CNN}. We also evaluate FINCH on MNIST raw pixel features with 784 (i.e., $28\times28$) dimensions, \texttt{MNIST\_8M\_PIXELS} and \texttt{MNIST\_70k\_PIXELS}.
\begin{table}[t]
\small
\tabcolsep=0.1cm
\begin{center}
\resizebox{7cm}{!} {
\begin{tabular}{l|c|c| c c c }
\toprule
&\multicolumn{2}{c}{ACC}  & P & R & F \\
\cline{2-6}   
Method & BFs05e02 & BBTs01e01  &  \multicolumn{3}{c}{Accio, \#C=40} \\
\midrule
ULDML~\cite{cinbis2011unsupervised}	& 41.62		& 57.00		& $-$ & $-$ & $-$\\
HMRF~\cite{wu2013constrained}		& 50.30		& 60.00 	&27.2 &12.8 &17.4\\
WBSLRR~\cite{xiao2014weighted}		& 62.76		& 72.00		&29.6 &15.3 &20.2\\
JFAC~\cite{zhang2016joint}			& {92.13}	& $-$	&71.1 &35.2 &47.1\\
Imp-Triplet~\cite{imptriplet}		& $-$		& 96.00	& $-$ & $-$ & $-$\\
VDF~\cite{sharma2017}		        & 87.46		& 89.62	& $-$ & $-$ & $-$\\
TSiam~\cite{sharma2019} 			& 92.46 & 98.58 & $76.3$ & $36.2$ & $49.1$ \\
SSiam~\cite{sharma2019} 			& 90.87 & 99.04 & $77.7$ & $37.1$ & $50.2$ \\
\midrule
\textbf{FINCH} 	& \textbf{92.73}       & \textbf{99.16} & \textbf{73.30}	 &\textbf{71.11}	&\textbf{72.19} \\
\bottomrule
\end{tabular}}
\end{center}
\vspace{-0.15cm}
\caption{SOTA comparison on face clustering.}\label{tab:soa}
\vspace{-0.5cm}
\end{table}

\subsection*{Face Clustering- State-of-the-art Comparison on BBTs01, BFs05 and Accio Datasets}
Here, first we report the pairwise F-measure performance on the ground-truth~(true) number of clusters (i.e. 5 for BBTs01~(season 1, episodes 1 to 6) and 6 for BFs05~(season 1, episodes 1 to 6)) using the steps described in Algorithm~2. FINCH obtains a pairwise F-measure of $\mathbf{97.42\%}$ and $\mathbf{94.02\%}$ respectively. In comparison, the reported pairwise F-measure of the recent work by Jin~\etal~\cite{erdosclustering} for BBTs01 and BFs05 are $78.2\%$ and $62.99\%$ respectively.

In addition to BFs05~(season 5, episodes 1 to 6) and BBTs01~(season 1, episodes 1 to 6), we also report the individual performances on the popularly used episodes of BFs05 and BBTs01, these are BFs05e02~(season 5, episodes 2) and  BBTs01e01~(season 1, episodes 1). BBTs01e01 has 41,220 frames, and BFs05e02 has 39,263 frames.  In Table~\ref{tab:soa}, we compare the performance of FINCH with the current state-of-the-art algorithms on BBTs01e01 and BFs05e02. FINCH estimates 7 clusters for BFs05e02, and is brought down to ground-truth number of 6 clusters using Algorithm~2, while for BBTs01e01 FINCH estimated exactly 5 clusters which is also the true number of clusters.


\begin{table*}[t]
\small
\tabcolsep=0.1cm
\begin{center}
\resizebox{18cm}{!} {
\begin{tabular}{  l | ccccccccccccc}
\toprule
Steps/Partitions&	MICE\_PROTEIN&	REUTERS&	HAR&	STL-10&	MNIST\_10k&	MNIST\_70K&	BBTs01&	BFs05&	MNIST\_8M\_CNN&	MNIST\_8M\_PIXELS\\

\midrule
\midrule
&	\textbf{1077}&	\textbf{10k}&	\textbf{10299}&	\textbf{13k}&	\textbf{10k}&	\textbf{70k}&	\textbf{199346}&	\textbf{206254}&	\textbf{8.1M}&	\textbf{8.1M}\\
\midrule
1&	351&	1837&	2465&	2061&	1699&	11493&	27294&	29150&	2015305&	1845149\\
-&	99.7214&	94.36&	96.7667&	95.9462&	99.25&	99.7243&	98.4053&	95.8561&	99.9999&	99.9999\\
\midrule
2&	106&	220&	369&	177&	310&	1891&	6067&	6830&	694525&	691696\\
-&	96.9359&	89.39&	92.776&	94.9846&	99.18&	99.68&	98.3285&	95.6137&	99.997&	99.9725\\
\midrule
3&	24&	24&	88&	37&	65&	357&	1406&	1753&	161371&	175426\\
-&	64.624&	82.2&	86.5521&	94.7846&	99.18&	99.6729&	98.2819&	95.0396&	99.9816&	98.6474\\
\midrule
4&	\textbf{8}&	\textbf{4}&	18&	\textbf{10}&	17&	75&	251&	355&	33217&	25609\\
-&	\textbf{51.2535}&	\textbf{66.14}&	74.4441&	\textbf{85.2846}&	99.18&	99.6729&	98.0918&	94.0888&	99.951&	94.3231\\
\midrule
5&	4&	-&	\textbf{6}&	2&	\textbf{10}&	17&	36&	50&	6945&	4085\\
-&	26.9266&	-&	\textbf{60.2389}&	20&	\textbf{99.18}&	99.6729&	97.9413&	94.0748&	99.9067&	93.1139\\
\midrule
6&	3&	-&	2&	-&	-&	\textbf{10}&	\textbf{6}& \textbf{7}&	1362&	789\\
-&	15.506&	-&	35.5957&	-&	-&	\textbf{99.6729}&	\textbf{97.9413}&	\textbf{94.0748}&	99.8833&	92.2973\\
\midrule
7&	-&	-&	-&	-&	-&	-&	2&	2&	262&	177\\
-&	-&	-&	-&	-&	-&	-&	49.7778&	54.6646&	99.8762&	90.3298\\
\midrule
8&	-&	-&	-&	-&	-&	-&	-&	-&	53&	43\\
-&	-&	-&	-&	-&	-&	-&	-&	-&	99.8762&	83.359\\
\midrule
9&	-&	-&	-&	-&	-&	-&	-&	-&	\textbf{13}&	\textbf{11}\\
-&	-&	-&	-&	-&	-&	-&	-&	-&	\textbf{99.8762}&	\textbf{66.6879}\\
\midrule
10&	-&	-&	-&	-&	-&	-&	-&	-&	4&	-\\
-&	-&	-&	-&	-&	-&	-&	-&	-&	41.1708&	-\\
\bottomrule
\end{tabular}}
\end{center}
\caption{{FINCH steps run for all used datasets. Total number of clusters in each partition along with its respective accuracy as measured by Clustering Accuracy~(ACC), each row is represented as ($_{ACC}^{\#clusters}$}).}  \label{table:partitions} 
\end{table*}

We also include FINCH results on Accio dataset~\cite{ghaleb2015accio} (“Harry Potter” movie series with a large number of dark scenes) with
36 named characters and 166885 faces/samples to cluster. The largest to smallest cluster ratios of Accio are very skewed: 30.65\% and 0.06\%. The performance on Accio is measured with B-Cube precision, recall and F-score on \#clusters=40 as in the compared methods.

Note that, FINCH has simply clustered the extracted VGG2 feature vectors for the frames, without any data specific feature training/transfer and without exploiting any form of video level constraints as used in the compared method: video-level constraints~\cite{cinbis2011unsupervised},  video editing style~\cite{tc}, dynamic clustering constraints in CNNs via MRF modeling~\cite{zhang2016joint}, triplet loss based CNN training using must-link and must-not-link constraints obtained from video-level constraints~\cite{imptriplet}, Siamese feature transfer/training~\cite{sharma2019} using track-level constraints (TSIAM) and a self-supervised transfer (SSIAM) on the extracted VGG2 feature vectors.

\subsection*{Partitions for each dataset}
Table~\ref{table:partitions} shows the total FINCH steps run for all used datasets. Each step produces a partition of data with shown number of clusters in each. The clustering accuracy~(ACC) at each step is reported that demonstrate the quality of the merges and the clusters obtained. As can be seen, despite the varying nature/distribution and dimensionality of data, FINCH is able to recover the ground-truth or a very close partition in all cases.

\subsection*{Details for baselines}
We have tried different parameters for the baselines, and where applicable used the recommended parameters (by respective authors), to report their best NMI scores. As an example, Table~\ref{table:AP_param} shows the impact of changing the preference parameter of Affinity prop (AP) on two datasets. As shown, on preference=-100 AP estimates ground truth 8 clusters on mice-protein data, but the performance is worst, while the same preference value on MNIST-10k data produces far more clusters than groundtruth while also lower NMI score. The hyper-parameters do not generalize on different data. This also motivates why a parameter-free clustering algorithm is needed.

For AP, SC, BR and kmeans++ we use the implementation available in the scikit-learn package. For RO, we use the implementation available in OpenBR framework \url{http://openbiometrics.org/}. For HAC we use the implementation provided by Matlab. For JP, we use the implementation available from~\cite{jp_matlab}. For RCC, we use the python implementation recommended by the authors. Similarly for SSC and MV-LRSSC we have used the Matlab implementations provided by the respective authors. Table~\ref{table:baselines} summarizes the used parameters for the baselines. 
\begin{table}[tb]
\small
\tabcolsep=0.2cm
\begin{center}
\resizebox{7.5cm}{!} {
\begin{tabular}{   l | cc c  c  | c | c  }
\toprule
    & \multicolumn{4}{c}{Affinity Propagation~(AP): Preference}  \\ 
\cline{2-5}
  Dataset   & Reported & -10 & -50 & -100 & FINCH & True \#C\\
\midrule
\midrule
Mice\_Protein   & \textbf{59.10} & 57.63 & 44.09 & 37.93 &\textbf{51.64} \\
Estim.~\#C      & 67 & 48 & 12 & \textbf{8} &\textbf{8} & 8 \\
\midrule
MNIST\_10k      & 69.97 & 58.02 & 63.97 & 67.13 &\textbf{97.55} \\
Estim.~\#C      & 116 & 1260 & 304 & 178 & \textbf{10} & 10\\
\bottomrule
\end{tabular}}
\end{center}
\vspace{-0.15cm}
\caption{Example: Impact of varying preference parameter of AP.}
\label{table:AP_param}
\end{table}
\begin{table}[tb]
\small
\tabcolsep=0.1cm
\begin{center}
\resizebox{8.5cm}{!} {
\begin{tabular}{   l  | l }
\toprule
    Algorithm &   Parameter settings \\
\midrule
\midrule
AP & Preference = median of similarities, damping factor = 0.5, max\_iter=200, convergence\_iter=15\\
JP & \# neighbors=10, min\_similarity\_to\_cluster =0.2,  min\_similarity\_as\_neighbor=0 \\
RO & distance threshold $t$=14,  number of top neighbors $K$=20 \\
RCC &  maximum total iteration $maxiter$=100, maximum inner iteration $inner\_iter$=4\\
BR & threshold=0.5, branching\_factor=50\\
Kmeans & init=``k-means++", n\_init=10, max\_iter=300\\
SC & eigen\_solver=\textit{``arpack"}, affinity=\textit{``nearest\_neighbors"}, n\_neighbors=10\\
SSC & r=0, affine=false, alpha=20, outlier=true, rho=1\\
MV-LRSSC & mu=0.0001, lambda1=0.9, lambda2=0.1, lambda3=0.7, noisy=true\\
\bottomrule
\end{tabular}}
\end{center}
\caption{Parameter settings for baselines} \label{table:baselines} 
\end{table}


\end{document}